\documentclass[letterpaper, 10 pt, conference]{ieeeconf}  

\IEEEoverridecommandlockouts                              

\usepackage{cite,comment}
\usepackage{url}            
\usepackage{booktabs}       
\usepackage{amsfonts}       
\usepackage{nicefrac}       
\usepackage{microtype}      
\usepackage{amsmath} 
\usepackage{cases}
\let\labelindent\relax
\usepackage{enumitem}

\usepackage{amsthm}
\theoremstyle{remark}
\newtheorem{remark}{Remark}
\usepackage{xcolor}
\usepackage{graphicx}
\usepackage{tikz}
\usetikzlibrary{decorations.pathreplacing}
\usepackage[utf8]{inputenc}
\usepackage{pgfplots}
\usepackage{xcolor}
\usepackage{subcaption} 
\usepackage{bm}
\usepackage{mathtools}
\usepackage{scalefnt}
\usepackage{dblfloatfix}
\allowdisplaybreaks

\usepackage{algorithm,algpseudocode}

\newcommand{\vmu}{\mbox{\boldmath $\mu$}}

\newcommand{\vom}{\mbox{\boldmath $\omega$}}

\newcommand{\vvph}{\mbox{\boldmath $\phi$}}

\newcommand{\vGa}{\bm \Gamma}

\newcommand{\vSi}{\bm \Sigma}

\newcommand{\vb}{\bm b}

\newcommand{\ve}{\bm e}
\newcommand{\vf}{\bm f}
\newcommand{\vg}{\bm g}
\newcommand{\vh}{\bm h}

\newcommand{\vq}{\bm q}

\newcommand{\vu}{\bm u}
\newcommand{\vv}{\bm v}

\newcommand{\vx}{\bm x}

\newcommand{\vz}{\bm z}

\newcommand{\vB}{\bm B}
\newcommand{\vC}{\bm C}

\newcommand{\vJ}{\bm J}
\newcommand{\vK}{\bm K}

\newcommand{\vM}{\bm M}

\newcommand{\vQ}{\bm Q}
\newcommand{\vR}{\bm R}
\newcommand{\vS}{\bm S}

\title{\LARGE \bf
Bayesian Multi-Task Learning MPC for Robotic Mobile Manipulation
}

\author{Elena Arcari$^{*,1}$, Maria Vittoria Minniti$^{*,2}$, Anna Scampicchio$^{1}$,\\ Andrea Carron$^{1}$, Farbod Farshidian$^{2}$,  Marco Hutter$^{2}$, and Melanie N. Zeilinger$^{1}$
\thanks{*The authors contributed equally to this work}
\thanks{$^{1}$These authors are with the Institute for Dynamic Systems and Control, ETH Zurich {\tt\scriptsize earcari|ascampicc|carrona|mzeilinger@ethz.ch}}%
\thanks{$^{2}$These authors are with the Robotic Systems Lab, ETH Zurich {\tt\scriptsize mminniti|farbodf|mahutter@ethz.ch}}%
\thanks{This work was supported by the Swiss National Science Foundation through the National Centre of Competence in Digital Fabrication (NCCR dfab), the Swiss National Science Foundation through the National Centre of Competence in Research Robotics (NCCR Robotics), and by Intel.} 
}

\begin{document}

\maketitle

\begin{abstract}
Mobile manipulation in robotics is challenging due to the need of solving many diverse tasks, such as opening a door or picking-and-placing an object. Typically, a basic first-principles system description of the robot is available, thus motivating the use of model-based controllers. However, the robot dynamics and its interaction with an object are aﬀected by uncertainty, limiting the controller's performance. To tackle this problem, we propose a Bayesian multi-task learning model that uses trigonometric basis functions to identify the error in the dynamics. In this way, data from diﬀerent but related tasks can be leveraged to provide a descriptive error model that can be efficiently updated online for new, unseen tasks. We combine this learning scheme with a model predictive controller, and extensively test the eﬀectiveness of the proposed approach, including comparisons with available baseline controllers. We present simulation tests with a ball-balancing robot, and door-opening hardware experiments with a quadrupedal manipulator.
\end{abstract}

\section{Introduction}
\label{sec:introduction}

In robotics, model uncertainties typically arise when the robot dynamics is not known precisely or when manipulating unknown objects. Uncertainty in model-based control has been treated in many ways. For instance, one can assume that modeling errors in the system dynamics belong to a bounded set, as in robust model predictive control (MPC). Alternatively, prior information in the form of a probability distribution may be leveraged in stochastic MPC, or Reinforcement Learning (RL) methods. While having achieved significant results in challenging non-linear problems, these formulations are designed to remain fixed during operation~\cite{LH20}. Planning for object manipulation, however, requires accurate knowledge of both the kinematic and dynamic parameters of the object. While the former can be identified through vision before executing the task \cite{mittal2021articulated}, the latter can only be determined by interacting with the object itself. This problem becomes even more challenging for dynamically-stable mobile manipulators, which require precise motion coordination of the mobile base and the arm while maintaining the balance. One way to tackle this is to refine the estimates of model errors using online adaptation or identification procedures as in, e.g., {learning-based MPC~\cite{LB22, sinha2022adaptive}}. \\ This paper addresses online model improvements in the context of generating high-performance MPC controllers for robotic mobile manipulation. While the MPC requires an uncertainty representation with low computational overhead, for online learning it is critical to have a descriptive error model that can be efficiently adapted with data available during operation. Therefore, the main challenge is to simultaneously address both issues despite not having necessarily seen the same task before. 
\begin{figure}[t]
   \centering
   \includegraphics[scale=0.22]{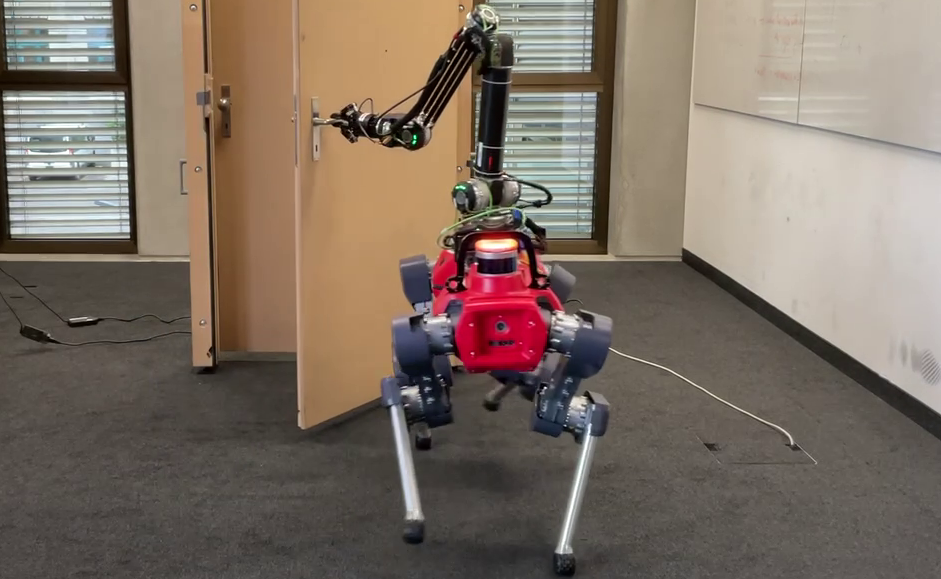}
   \caption{The quadrupedal robot ANYmal opening a door. The proposed method learns the model of the manipulated object using data from multiple related tasks (e.g. doors of different masses). When interacting with a new object, the learned model is adapted online using data collected during operation.
}
\label{fig:title_page}    
\end{figure}
\\Robotic manipulation tasks often share a common structure:
for example, when manipulating intra-category objects, even though some physical parameters such as damping and stiffness may vary, the model structure remains mostly the same. We leverage this by proposing a Bayesian multi-task, learning-based MPC scheme, where the model error is approximated by means of shared trigonometric basis functions, which are linearly combined with task-specific parameters~\cite{arcari2021bayesian}. The choice of sinusoidal features provides a low-complexity representation of the error dynamics~\cite{MLG10}, reducing the computational burden of the MPC in evaluating the model and its derivative along the time-horizon. The proposed scheme consists of a training and an online learning phase. In the training phase, we optimize over shared error model hyperparameters using a set of data acquired while performing similar tasks. 
In the online learning phase, we fix the previously identified basis functions, and fine-tune the error model by adapting the linear parameters using data collected during a new, and potentially unseen, operation. 
{\subsubsection*{Contributions}
The contributions of this letter are the following. First, we combine MPC with an adaptive, multi-task model learning method, which is based on a parametric sinusoidal feature representation~\cite{arcari2021bayesian}. The proposed model enables the scheme to run in real-time on a robot's onboard computer, while identifying the error in the MPC system dynamics. \\
Secondly, we present the application of the proposed method to robotic mobile manipulation problems. We test the controller on (i) an underactuated ball-balancing robot (\textit{ballbot})\cite{minniti2019whole} in the presence of uncertainty in its dynamics, and (ii) a quadrupedal manipulator \cite{sleiman2021unified} in a door-opening task with a partially unknown door model (see Fig.~\ref{fig:title_page}). \\ In addition, we present extensive simulation and hardware results, comparing the proposed approach against other adaptive MPC controllers where the model error is identified via single-task learning \cite{MLG10} and using non-sinusoidal basis functions, in particular neural networks \cite{JH18b}.} 
\subsubsection*{Related Work}
Multi-task learning, also known as \emph{learning-to-learn} or \emph{meta-learning}, leverages past data in the form of previously observed tasks, to construct an initial prior that can be efficiently used to learn in unseen scenarios. For instance, in~\cite{KP18}, the authors propose a multi-robot transfer learning framework that combines $\mathcal{L}_1$-adaptive control and iterative learning control. In~\cite{RS21}, control-oriented meta-learning is used to identify the non-linear basis functions of the employed adaptive controller. While addressing the issue of efficient online adaptation, these controllers do not exploit the inherent constraint handling and predictive capability of MPC. \\
In model-free RL, learning from multiple tasks was employed to mitigate data demand, as e.g. in~\cite{AM18, ZX19, song2020rapidly, ghadirzadeh2021bayesian}. However, these approaches generally resort to sim-to-real transfer learning~\cite{yu2020learning}, and therefore require very accurate models of real-world systems, which are often imprecise or unavailable. \\
The combination of single-task learning and model-based control has been successfully applied to multiple robotic platforms, e.g. for fine-tuning nominal models with a learned residual~\cite{AC19, LH19}, for complex manipulation tasks~\cite{lenz2015deepmpc, minniti2021model}, or to adapt to the presence of unknown loads~\cite{sun2021online, minniti2021adaptive, hanover2021performance}. However, typically past information from previously observed tasks is not leveraged in a structured manner. This can be enabled, for instance, via multi-task learning, whose combination with model-based approaches has been studied, e.g. in~\cite{LZ20, IC19}. Multi-task learning methods often make use of non-parametric techniques, as in~\cite{SS18}, which exploits latent variable Gaussian processes, or {in~\cite{SB21, lew2022safe, MC22}, where neural networks are used}. While these frameworks can effectively learn complex non-linear dynamics, applying them to high-dimensional robotic systems, where stringent computational requirements are imposed by practical hardware limitations, may be challenging.
The low-complexity option that we pursue in this work makes use of parametric models, in particular linear combinations of sinusoidal basis functions~\cite{arcari2021bayesian}. Similar ideas are developed in~\cite{CM21}, and in~\cite{JH18a, lew2022safe}, which employ Bayesian last-layer neural networks. 
\section{Multi-task learning for MPC}
\label{sec:multi-task learning MPC}
\begin{figure*}[t]
   \centering
   \includegraphics[scale = 0.52]{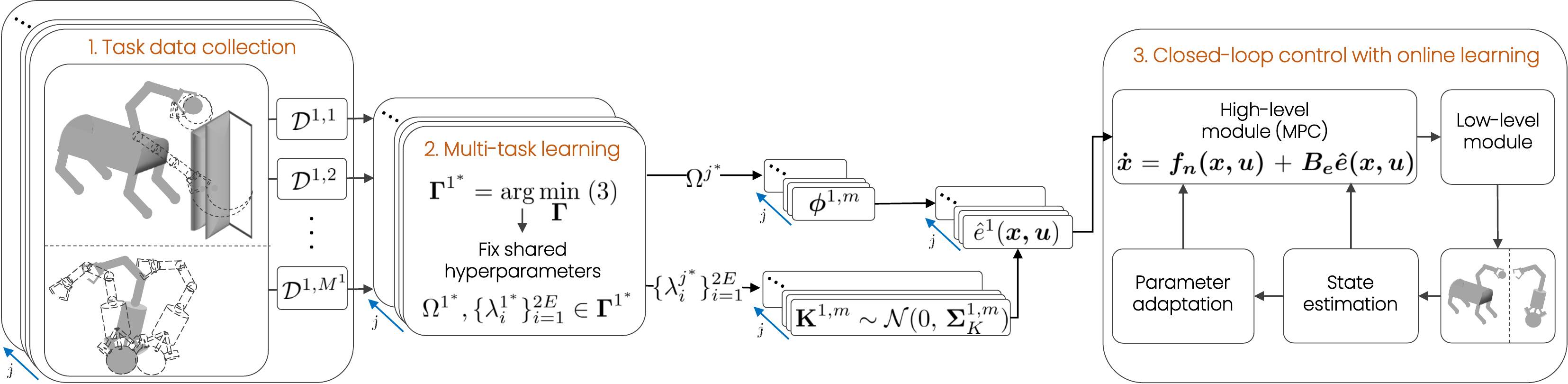}
   \caption{Overall pipeline of the proposed multi-task learning MPC method for each element $j = 1,\dots,n_e$ of the error vector. For data collection, we let the robot perform $M$ similar tasks under the nominal MPC controller (block 1). This training data is used to optimize a fixed set of shared hyperparameters (block 2). During online execution, the linear parameters $\{\vK\}_{j=1}^{n_e}$ are updated via Kalman filtering, and the modeling error is compensated in the MPC (block 3).} 
\label{fig:pipeline}    
\end{figure*}

We consider the following non-linear, continuous-time system dynamics
\begin{equation}
    \dot \vx = \vf_{n}(\vx, \vu) + \vB_e \, \ve(\vx, \vu) , \label{eq:system_dynamics}
\end{equation}
where $\vx(t) \in \mathcal{X} \subseteq \mathbb{R}^n$, $\vu(t) \in \mathcal{U}\subseteq \mathbb{R}^m$ are the state and input of the system. The dynamics of Eq.~\eqref{eq:system_dynamics} is modeled as the sum of a \textit{nominal} component $\vf_{n}$, and an error term $\ve$, which captures the unknown part of the system and affects $n_e$ components of the state selected by 
$\vB_e \in \mathbb{R}^{n \times n_e}$. For a robot, this error is typically due to an inaccurate knowledge of the system dynamics parameters, such as masses, center of masses (CoM), or inertias of each link, or due to interaction with the external environment. {In this paper, the error model $\ve$ is approximated as $\hat\ve$, where each vector component $\hat e^i, i \in \{1, \dots, n_e \}$ is defined as 
\begin{equation}
e^i(\vx, \vu) \approx \hat e^i(\vx, \vu) = \vvph^i(\vx, \vu)^\top \vK^i, 
\label{eq:approx_error_model}
\end{equation}
where $\vK^i$ is the vector of \textit{task-specific} parameters that linearly combines the \textit{shared} trigonometric basis functions contained in the vector $\vvph^i(\vx,\vu)$. The results of this letter show that the inherent nonlinearity of Eq.~\eqref{eq:approx_error_model} provides a sufficiently rich description, and at the same time offers a simple structure allowing for efficient computation and online adaptation.} \\
{In the following, we first describe the framework for learning the parametric approximation in Eq.~\eqref{eq:approx_error_model} using a multi-task training procedure based on~\cite{arcari2021bayesian}. Specifically, we employ Kullback-Leibler divergence minimization to optimize the hyperparameters that are shared among tasks, i.e., the frequencies of the trigonometric basis functions $\vvph^i(\vx,\vu)$.}
Additionally, we discuss the online learning phase, during which the basis functions of the approximate parametric error model are fixed, and the task-specific parameters combining them are adapted online by means of a Kalman filter estimation scheme. Finally, we introduce the optimal control problem (OCP) formulation, and the employed receding horizon strategy generating real-time solutions. A summary of the overall methodology is provided in Fig.~\ref{fig:pipeline}.


\subsection{Training phase: learning from multiple tasks}
\label{sec:Training phase: learning from multiple tasks}

{We consider the collection of data from $M$ tasks, i.e. noisy error measurements gathered under different operating conditions. Denoting with $\mathcal{J}$ the set of $n_e$ indices of the state vector 
affected by $\ve(\vx, \vu)$ in \eqref{eq:system_dynamics},
the measurement model for each $j \in \mathcal{J}$ is
%
\begin{equation}
    y^{j,m} = e^{j,m}(\vx^{m}, \vu^{m}) + w^{j,m}, \quad m = 1, \dots, M, \notag
\end{equation}
where $w^{j,m}  \sim \mathcal{N}(0, \sigma_w^{j,m})$ is zero-mean Gaussian, task-specific noise with variance $\sigma_w^{j,m}$.
The error $e^{j,m} = \dot{x}^{j,m} - f^j_n(\vx^{m}, \vu^{m})$ is computed as the difference between the $j$-th component of the state derivative and the nominal flow-map, evaluated at the \textit{input location}~$(\vx^{m},\vu^{m})$.} In the remainder of the section, we drop the index $j$ for ease of notation and refer to a generic vector element.

\subsubsection{Trigonometric basis functions error model}
\label{subsec:Trigonometric_basis_functions_model}
We store the data collected for each task $m$ into the set $\mathcal{D}^m = \{ (\vx^m_i, \vu^m_i) , y^m_i \}_{i=1}^{N_m}$ with cardinality $N_m$. The approximate error model $\hat{e}^m(\cdot, \cdot)$, evaluated at each state-input pair in $\mathcal{D}^m$, is defined as
\begin{equation}
    \hat{e}^{m}(\vx^m_i, \vu^m_i) = \vvph^m(\vx^m_i, \vu^m_i)^\top \vK^m, \quad i = 1 \dots N_m,
    \notag
\end{equation}
where $\vK^m \in \mathbb{R}^{2E}$ is the vector of task-specific parameters. The entries of $\vvph^m(\cdot.\cdot)$ are defined for each $i = 1, \dots, N_m$ as
\begin{equation}
[\vvph^m(\vz_i^m)]_{l} = \begin{cases}
        \sin(2\pi \vom_l^\top \vz_i^m)         \hspace{2.5em} l = 1, \dots, E\\
        \cos(2\pi \vom_{l - E}^\top \vz_i^m) \hspace{1.5em} l = E + 1, \dots, 2E,
    \end{cases}
    \notag
\end{equation}
where $\vz^m_i$ is the vector concatenating the input locations $\vx_i^m$ and $\vu_i^m$, and $E$ is the pre-specified number of shared frequencies contained in a set $\Omega = \{\vom_1, \dots, \vom_{E} \}$. \\ We assume that the vector of coefficients $\vK^m$ has a Gaussian prior distribution, such that $\vK^m \sim \mathcal{N}(0, \vSi_K)$, with $\vSi_K = \text{diag}([\lambda_1, \dots , \lambda_{2E}])$. {All hyperparameters $\{ \lambda_i \}_{i=1}^{2E}$, $\{ \sigma_w^m \}_{m=1}^M$, and $\Omega$ are collected in a vector $\vGa$, such that the posterior model for the $m$-th task is expressed as
\begin{equation}
    \hat{e}^{m}(\vx^m_i, \vu^m_i)_{| y^m_i, \vGa} =  \vvph^m(\vx^m_i, \vu^m_i)^\top \hat\vK^m, \quad i = 1 \dots N_m,
    \notag
\end{equation}
where $\hat\vK^m \sim \mathcal{N} (\hat{\vmu}_K, \hat{\vSi}_K)$ is obtained via Bayesian inference, exploiting the fact that the Gaussian distribution is self-conjugate.}

\begin{remark}
Note that this setup can be described as in~\cite{arcari2021bayesian} by considering a shared set of input locations $\bigcup_{m=1}^M \{\vx^m_i, \vu^m_i \}_{i=1}^{N_m}$. In this case, the noise variance acts as a selector by specifying it as both task and input specific, i.e., for $i = 1, \dots, N_m$:
\begin{equation*}
\sigma_{w,i}^{m} = \begin{cases}
        \sigma_w^{m}         \hspace{3em} \text{if } y^{m}_i = e^{m}(\vx^m_i, \vu^m_i) + w^{m}_i \text{ exists}\\
        \infty \hspace{3.2em} \text{otherwise}.
    \end{cases}
\end{equation*}
\end{remark}

\subsubsection{Hyperparameter optimization problem}

We leverage data from $\{\mathcal{D}^m\}_{m=1}^M$ for optimizing the vector of hyperparameters $\vGa$: for each task, a posterior predictive distribution $q^m(\cdot)$ is associated with the approximate model described in Section~\ref{subsec:Trigonometric_basis_functions_model}, and its ``distance" to the true unknown task predictive distribution $p^m(\cdot)$, expressed in terms of the Kullback-Leibler (KL) divergence, is minimized with respect to $\vGa$~\cite{arcari2021bayesian}. {The KL divergence is originally defined as the sum of expectations of $\log(q^m(\cdot))$, taken with respect to the unknown $p^m(\cdot)$. To express these distributions, the available dataset $\mathcal{D}^m$ is split into two parts, i.e.  $\mathcal{D}^m_t$ with cardinality $N_t$, and $\mathcal{D}^m_v$ with cardinality $N_v$. The set $\mathcal{D}^m_t$ is used to compute the posterior distribution of the vector $\hat\vK^m$, which in turn enables the inference of $q^m(\cdot)$. The data in $\mathcal{D}^m_v$ is used to approximate the expected value as an averaged sum, i.e. evaluating the predictive distribution $q^m(\cdot)$ for all input locations $(\bar\vx ,\bar\vu ) \in \mathcal{D}^m_v$. The final optimization problem is}
\begin{equation}
    \min_{\vGa} -\frac{1}{M} \sum_{m=1}^M \frac{1}{N_v} \sum_{ (\bar\vx,\bar\vu) \in \mathcal{D}^m_v } \log q^m(\bar y | (\bar\vx,\bar\vu), \mathcal{D}^m_t, \vGa ) ,
    \label{eq:multi-task learning}
\end{equation}
with optimizer $\vGa^* = [ \Omega^*, \{ \lambda^*_i \}_{i=1}^{2E}, \{{\sigma^{m}_w}^* \}_{m=1}^M]$. \\
{Minimizing the KL divergence corresponds to a maximum likelihood estimation of the hyperparameters $\vGa$ of the approximate predictive distribution $q^m(\cdot)$. Furthermore, thanks to the employed Bayesian framework, all involved distributions are Gaussian, and therefore Eq.~\eqref{eq:multi-task learning} can be expressed in terms of mean and covariance of $q^m(\cdot)$.}
\subsection{Online learning phase: task-specific parameter adaptation}
\label{sec:adaptation_to_unseen_tasks}

Once the trigonometric basis functions' frequencies $\Omega^*$ are inferred during the training phase, they are kept fixed during a new control task, which in general is not one of the $M$ tasks previously observed during training. {The vector of task-specific parameters~$\vK^j_0$ is initialized at its optimized prior distribution~$\mathcal{N}(0, \vSi^{j^*}_K)$ and, following the proposed Bayesian framework, is adapted via Kalman filtering using measurement data collected online.} {The parameter process and measurement models are defined as 
\begin{subequations}
\begin{align}
    & \vK^j_{k+1} = \vK^j_{k} + v^j_k, \quad \qquad \qquad j = 1, \dots, n_e \label{eq:process model}\\
    & y^j_{k} = \vvph^j(\vx_k, \vu_k)^\top \vK^j_k +  w^j_k, \quad j = 1, \dots, n_e \label{eq:measurement model}
\end{align}
\label{eq:Kalman filter for parameters}
\end{subequations}
where~$v^j_k \sim \mathcal{N}(0,\sigma^j_Q I_{2E})$ is i.i.d. Gaussian process noise, with~$\sigma^j_Q, I_{2E}$ being the variance of each element of the vector $\vK^j_{k}$ and the identity matrix of dimension~$2E$, respectively, and~$w^j_k \sim \mathcal{N}(0,\sigma^j_R)$ is i.i.d. Gaussian measurement noise with variance $\sigma^j_R$. }
As specified in Section~\ref{subsec:Trigonometric_basis_functions_model}, the measurement noise is considered to be task-specific. For a new, unseen task, one can estimate its variance $\sigma^j_R$ online via marginal likelihood optimization with an initial batch of data~\cite{CR17},~\cite{MLG10} prior to adapting~$\vK^j$. \\ Note that the performance of the Kalman filter in Eq.~\eqref{eq:Kalman filter for parameters} crucially depends on~$\sigma_Q^j$, and~$\sigma_R^j$, and on the accuracy of the error model in Eq.~\eqref{eq:measurement model}. In particular, if~\eqref{eq:measurement model} represents the true model, and under further assumptions on~$\vvph^j$, a tracking error bound for $\vK^j_k$ is available~\cite{LG88}. Even if these assumptions are not satisfied, it is possible to compensate for poor knowledge of the error model by manually increasing the filter's confidence on data, e.g. by increasing~$\sigma^j_Q$, or by decreasing~$\sigma^j_R$.
\subsection{MPC formulation}
\label{sec:mpc_formulation}

 The control problem associated with the system described by Eq.~\eqref{eq:system_dynamics} is formulated as the following non-linear OCP:
\begin{subequations}
\begin{align}
     & \underset{\vu(\cdot)}{\text{minimize}} & & l_f(\vx(T)) + \int_0^T l(\vx(t), \vu(t), t) \, dt \label{eq:cost_function} \\
    &\text{subject to:}  &&\dot \vx = \vf_{n}(\vx, \vu) + \vB_e \, \ve(\vx, \vu) \label{eq:system_dynamics_mpc}\\
    & & & \vx(0) = \vx_{0} \label{eq:initial_conditions} \\
    & & &\vg(\vx, \vu) = \mathbf{0}  
    \label{eq:state_input_equality_constraints} \\
    & & &\vh(\vx, \vu) \geq \mathbf{0}, \label{eq:inequality_constraints}
\end{align}
\label{eq:ocp}
\end{subequations}
where $l(\cdot,\cdot)$ is the stage cost, $l_f(\cdot)$ is the terminal cost, and~$T$ is the time-horizon. 
Eq.~\eqref{eq:initial_conditions} denotes the initial conditions of the problem, while $\vg$ and $\vh$ denote state-input equality and inequality constraints, respectively. The model error $\ve$ is approximated as $\hat{\ve}$ following the method presented in Sec.~\ref{sec:Training phase: learning from multiple tasks}, and its parameters $\{\vK^j\}_{j=1}^{n_e}$ are adapted online as discussed in Sec.~\ref{sec:adaptation_to_unseen_tasks}. \\
In this work, we employ an MPC scheme where, at each time-step, the OCP in \eqref{eq:ocp} is transformed into a finite-dimensional non-linear program (NLP) using a direct multiple-shooting approach \cite{bock1984multiple}. The resulting NLP is solved in a receding horizon fashion using Sequential Quadratic Programming (SQP), for which a detailed description 
is given in \cite{grandia2022perceptive}. The MPC problem is solved in a real-time iteration scheme \cite{diehl2005real}, where only one SQP step is performed per MPC update.\\
Here, physical limits~\eqref{eq:state_input_equality_constraints},~\eqref{eq:inequality_constraints} are implemented as soft constraints in order to preserve feasibility of the MPC problem~\eqref{eq:ocp}. Under further assumptions~\cite{LB22}, stronger guarantees in terms of closed-loop constraint satisfaction in probability can be provided by augmenting the control framework with safety certification mechanisms~\cite{TL22},~\cite{BT21}.

\section{Mobile manipulators}
\label{sec:robotic_mobile_manipulators}

{In this section, we describe how to tailor the OCP in~\eqref{eq:ocp} to two application scenarios: the motion planning problem of a ball-balancing manipulator (\textit{ballbot}), and the manipulation planning problem of a quadrupedal manipulator. 

\subsubsection{Ball-balancing manipulator}
\label{subsec:ballbot_method}
We consider a ballbot  with a 3-DoF arm mounted on top (see Fig.~\ref{fig:pipeline})~\cite{minniti2019whole}. We model the motion of the robot with $8$ degrees of freedom (DoFs), which are the planar position of the ball, the ZYX angles for the base orientation, and the arm joint angles. Let $\vx:=(\vq, \dot \vq)$, where $\vq \in \mathbb{R}^8$ are the robot generalized coordinates. 
The system dynamics of Eq.~\eqref{eq:system_dynamics} is then obtained as
\begin{equation}
\small
\label{eq:ballbot_eom}
    \dot \vx = \begin{bmatrix} \dot \vq \\
    \vM(\vq)^{-1}[-\vC(\vq, \dot \vq)\dot \vq -\vg(\vq)+ \vS(\vq)\vu] +\ve(\vq, \dot \vq)
    \end{bmatrix},
\end{equation}
where $\vM\in\mathbb{R}^{8 \times 8}, \vC \in \mathbb{R}^{8 \times 8}, \vg \in \mathbb{R}^{8}, $ are the nominal mass matrix, Coriolis matrix, and vector of gravitational terms; $\vS\in\mathbb{R}^{8\times 6}$ is an actuator-selector matrix, and $\vu\in \mathbb{R}^6$ corresponds to the robot actuation torques.\\ The vector $\ve \in \mathbb{R}^{8}$ lumps the uncertainties due to inaccurate knowledge of the robot dynamic parameters (such as masses, CoMs and inertias), which cause the mass matrix, Coriolis matrix and gravitational terms of the real system to deviate from the nominal terms in Eq.~\eqref{eq:ballbot_eom}. 
We note that, even though uncertainties in the mass matrix can cause the error $\ve$ to depend on the input $\vu$, in our implementation we neglect this dependence and assume that $\ve$ is only a function of the state, which allows for reducing the dimensionality of the problem. 
\subsubsection{Quadrupedal manipulator}
 \label{sec:object_manipulation}
The planning problem for a quadrupedal robot manipulating an object (Fig.~\ref{fig:title_page}) was originally presented in \cite{sleiman2021unified}. Here, we focus on the part of the OCP~\eqref{eq:ocp}  that concerns the robot interaction with an external object, and extend it to the case where the object dynamics is partially unknown. 
With reference to \cite{sleiman2021unified}, let $\vx_r, \vu_r$ be the robot state and input, respectively.
We assume that the robot is manipulating an object at its end-effector, and define $\vx_o = (\vq_o, \vv_o) \in \mathbb{R}^{2n_o}$ as the object state, where $\vq_o, \vv_o$ are the object generalized positions and velocities, respectively. The equations of motion of the object can be obtained as
\begin{equation}
    \dot \vx_o = \begin{bmatrix} \vv_o \\
    \vM_o^{-1}(-\vJ_{ee}^T\vf_{ee}-\vb_o)+\ve(\vq_o, \vv_o)
    \end{bmatrix},
    \label{eq:object_eom}
\end{equation}
where $\vM_o, \vb_o, \vJ_{ee}, \vf_{ee}$ are the object mass matrix, the vector of Coriolis, centrifugal, and gravitational terms, the Jacobian of the end-effector contact point, and the force exerted by the object on the robot at the end-effector, respectively. Furthermore, the error term $\ve \in \mathbb{R}^{n_o}$ captures all the uncertainties in the object dynamic properties. Crucially, $\ve$ is often not known prior to robot-object interaction, and therefore usually neglected.
Letting $\vx:=(\vx_r, \vx_o)$, $\vu = (\vu_r, \vf_{ee})$, we specify the system dynamics of the MPC problem in Eq.~\eqref{eq:system_dynamics}. \\ The stage cost $l$ for the MPC problem in Eq.~\eqref{eq:cost_function} is defined as $l=l_r+l_o$. The term $l_r$ penalizes robot-dependent quantities, and $l_o$ is a tracking cost given by
\begin{equation}
    l_o = \frac{1}{2}(\vx_o - \vx_o^{des})^T\vQ_o(\vx_o-\vx_o^{des}) + \frac{1}{2}\vf_{ee}^T\vR_{ee}\vf_{ee},
    \label{eq:object_cost}
\end{equation}
where $\vx_o^{des}$ is a vector of desired object generalized coordinates, and $\vQ_o\in\mathbb{R}^{n_o\times n_o}$, $\vR_{ee}\in\mathbb{R}^{3\times 3}$ are weight matrices.}

\section{Results}
\label{sec:results}

The following section is devoted to the empirical evaluation of the proposed approach. First, an object tracking scenario with a simulated quadrupedal manipulator is considered in Sec.~\ref{subsec:object_tracking_under_disturbance}. Further simulations with the ballbot are presented in Sec.~\ref{subsec:ballbot}, where the robot needs to follow end-effector targets while keeping balance despite model perturbations. In Sec.~\ref{subsec:door_opening_results}, experimental validation is provided in a door-opening task with a quadrupedal manipulator. A video showcasing the results accompanies this letter\footnote{https://youtu.be/IvlemJKH1aA}. \\ The following baseline MPC controllers are compared:
\begin{enumerate}[start=1,label={B\arabic*)}]
    \item \textit{nominal MPC}: the controller plans with the \textit{nominal} model and neglects the plant-model mismatch~\cite{sleiman2021unified},~\cite{minniti2019whole}.
    \item \textit{ground-truth MPC}: the controller uses the \textit{ground-truth} model of the system {(simulations only)}. 
    \item \textit{offset compensation}: the error in the MPC dynamics is approximated with an adaptive constant bias, similar to offset-free MPC~\cite{GP03}. 
    \item \textit{single-task MPC}: the basis functions hyperparameters are trained on a single dataset. The regression model has the same structure as proposed, and the hyperparameter optimization is based on~\cite{MLG10}.
    \item \textit{multi-task MPC}: the proposed method.
    \item \textit{neural network MPC}: neural network basis functions are considered, similar to \cite{JH18b}, and all hyperparameters involved are optimized following Eq.~\eqref{eq:multi-task learning}. 
\end{enumerate}

Note that the approach B2 is a theoretical baseline that can only be implemented in a numerical simulation environment, {and that neither B1 nor B2 perform any online adaptation.} When testing B3, B4, B5 and B6, the process and measurement noise variances (see Sec.~\ref{sec:adaptation_to_unseen_tasks}) have the same tuning in each examined scenario. {In Sections~\ref{subsec:object_tracking_under_disturbance} and \ref{subsec:ballbot}, we choose a very low process noise and fix the measurement noise to the true value as this allows for isolating the effect of the specific type of features (i.e. constant, trigonometric, or neural network basis functions).} Furthermore, the number of frequencies for B4 and B5 is fixed to $E = 3$, which we observed to work well for all proposed benchmarks. In Eq.~\eqref{eq:multi-task learning}, we ensure that both training and validation datasets contain measurements of a set-point change. \\
In all simulations and experiments, we use the OCS2 toolbox \cite{OCS2} for the underlying MPC computations. The hyperparameter optimization for multi-task MPC is carried out with Ipopt~\cite{AW06} within  the  CasADI~\cite{JA19} environment, while for neural network MPC it is implemented in PyTorch and solved with Adam~\cite{PD15}.
\subsection{Quadruped: object tracking under disturbance}
\label{subsec:object_tracking_under_disturbance}
{
We simulate a quadrupedal manipulator moving a 1 kg load at its end-effector along specified references. 
The object state $\vx_o$ is given by the position $[x_o, y_o, z_o]$ and velocity $[\dot x_o, \dot y_o, \dot z_o]$ of the load. The equations of motion of the object (Eq.~\eqref{eq:object_eom}) are artificially perturbed, and we assume that the disturbance on the load is only acting along the world $z$ direction. We first study the case of a sinusoidal disturbance and analyse the performance of multi-task MPC against baselines B1-4. Afterwards, we consider a  more complex nonlinear disturbance and compare against baseline B6. 
\subsubsection{Sinusoidal disturbance}
Each training task is defined by the particular unknown disturbance affecting the trajectory tracking, i.e. Task 1 is determined by a constant disturbance~$e_1(z_o) = -1.6$, while Tasks 2 and 3 by a sinusoid~$e_i(z_o) = 3\sin (2\pi \bar\omega_i z_o), \; i=2,3$, with~$\bar\omega_2 = 2$, and~$\bar\omega_3 = 6$. Task 4, encountered only during operation, is again induced by a sinusoidal disturbance~$e_4(z_o) = 3\sin (2\pi \bar\omega_4 z_o)$ with~$\bar\omega_4 = 4$. 
Simulation results, summarized in Tab.~\ref{table:alma_sims}, are given in terms of median closed-loop cost regret, which is the difference between the closed-loop costs of the ground-truth MPC and each considered baseline. The closed-loop costs are given by Eq.~\eqref{eq:object_cost}. Observing Tab.~\ref{table:alma_sims}, for Task 1 all methods outperform the nominal MPC and show comparable performance, while for Task 2 the proposed multi-task approach outperforms all baselines. The single-task MPC (3), i.e. trained on that particular task, achieves the best performance on Task 3. Finally, despite having higher median cost regret, for Task 4 the multi-task MPC demonstrates superior generalization capabilities since the cost regret of both single-task MPC (1) and offset compensation belong to the range~$[0.1397, 0.3443]$, while the multi-task MPC cost regret ranges in~$[0.1434, 0.1603]$. 
\begin{table}
\centering
\caption{Median cost regret w.r.t. ground-truth MPC. Simulation results for the nominal MPC, single-task MPC (using each training dataset), offset compensation, and multi-task MPC 
are averaged over 100 noise realizations.}
\setlength{\tabcolsep}{5pt}
\begin{tabular}{lcccc}
\hline
  & \multicolumn{1}{c}{Task 1} & \multicolumn{1}{c}{Task 2} & \multicolumn{1}{c}{Task 3} & \multicolumn{1}{c}{Task 4} \\ \hline
nominal MPC  & 0.3029 & 0.2341 & 0.3730 & 0.5517 \\ \hline
single-task MPC (1)  & 0.0049 & - & - & 0.1477 \\ \hline
single-task MPC (2)  & - & 0.1100 & - &  0.1681 \\ \hline
single-task MPC (3)  & - & - & \textbf{0.2412} & 0.1956 \\ \hline
offset compensation  & 0.0049 & 0.1013 & 0.4328 & \textbf{0.1476} \\ \hline
multi-task MPC  & \textbf{0.0047} & \textbf{0.0401} & 0.3313 & 0.1546 \\ \hline
\end{tabular}
\label{table:alma_sims}
\end{table}
\subsubsection{Sinusoidal and exponential disturbance}
Here we define the disturbance as $e(z_0) = a \sin(-2 \pi \bar\omega z_0) + b e^{[z_0, \dot{z}_0] k} $, where $\bar\omega = -3$ and $k = [1, 1]^\top$. To generate different tasks, we vary $ a \in [-6, 6]$, $b \in [-8 , 8]$, and pick 15 combinations as training tasks, while we use $(a,b) = (6,-8)$ as test task. 
For the neural network MPC, we choose a multi-layer perceptron composed of two hidden layers, each with \textit{tanh} activation function, and combined with a last linear $L$-dimensional layer, i.e., $L$ features. We perform an ablation study of the neural network model by varying the number of training tasks in $\{5, 10, 15\}$, the number of neurons in each hidden layer in $\{20, 50, 100 \}$, and $L$ in $\{9, 20\}$. We perform a similar study for the proposed multi-task model, choosing the number of training tasks in $\{ 3, 5 ,7\}$, and the number of features in $\{ 6, 9\}$. In Tab.~\ref{table:mtvsnn}, we report the best cost regret attained on the test task for a fixed number of features. We observe that, with the same number of features, multi-task MPC and neural network MPC perform comparably, and increasing the dimension of the neural network output layer does not improve the result. 

Overall, the multi-task MPC shows the most consistent behavior and improved generalization capabilities compared to all single-task and offset compensation counterparts. Furthermore, we demonstrate that the proposed approach, even if fairly simple and easy to tune, is both computationally attractive and capable of handling non-trivial disturbances, while the neural network MPC requires more critical design choices.
\begin{table}[h!]
\centering
\caption{Median cost regret w.r.t. ground-truth MPC. Simulation results for multi-task MPC using 6 and 9 features, trained on 5 tasks, and neural network MPC using 9 and 20 features, trained on 10 tasks, are averaged over 100 noise realizations.}
\setlength{\tabcolsep}{5pt}
\begin{tabular}{lrrr}
\hline
Method & \multicolumn{1}{c}{Median} & \multicolumn{1}{c}{$75\%$-quantile} & \multicolumn{1}{c}{$25\%$-quantile} \\ \hline
nominal MPC  & 4.1716 & - & - \\ \hline
multi-task MPC (6)  & 0.6951 & 0.7281 & 0.3620  \\ \hline
\textbf{multi-task MPC (9)}  & \textbf{0.0212} & \textbf{0.0189} & \textbf{0.0240}  \\ \hline
neural network MPC (9)  & 0.1974 & 0.1981 & 0.1969 \\ \hline
neural network MPC  (20) & 0.1901 & 0.1915 & 0.1890  \\ \hline
\end{tabular}
\label{table:mtvsnn}
\end{table}  
}
\subsection{Ballbot: perturbing the dynamics model}
\label{subsec:ballbot}

\begin{figure}[t]
   \centering
   \includegraphics[scale=0.23]{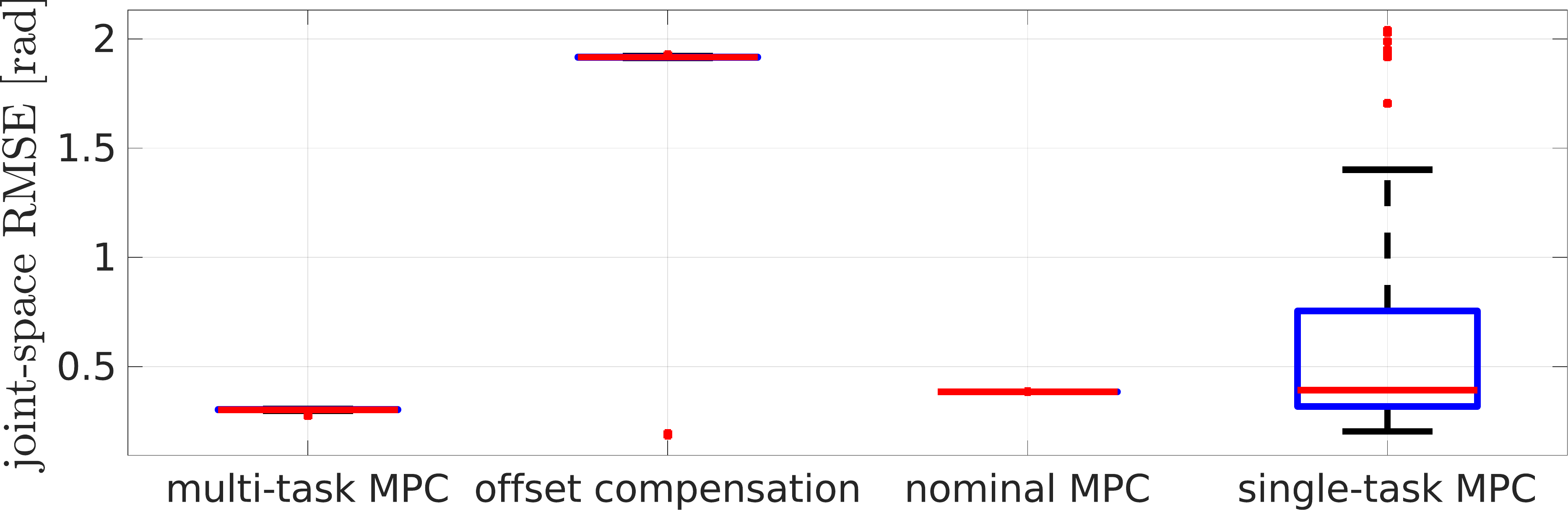}
   \caption{Boxplot showing the RMSE computed over 100 noise realizations between the ballbot closed-loop joint angles under the ground-truth MPC controller and the baseline controllers.} 
\label{fig:joint_space_error_boxplot_ballbot}    
\end{figure}

In the following simulations, the parameters in the ballbot dynamics are perturbed by adding a 1 cm offset along the y axis of the base CoM, a 1 mm offset along all the components of the CoM of the second and last link of the arm, and a 0.355 kg offset to the mass of the last link of the arm. As shown in the accompanying video, this perturbation is enough to cause the closed-loop state under the nominal MPC controller to considerably deviate from the ground-truth state. \\
The data-collection procedure consists in driving the robot along 7 different trajectories, such that each trajectory, designed to excite a particular degree of freedom, determines a task. The baselines B1-B5 from Sec.~\ref{sec:results} are tested on a new, unseen trajectory that can be visualized in the accompanying video. For all methods, the error $\ve$ in Eq.~\eqref{eq:ballbot_eom} is chosen as a function of the full-state of the robot excluding the planar position and the orientation around the world vertical axis. Note that for the single-task MPC, we randomly pick one of the available task datasets for hyperparameter training. The employed MPC time-horizon is 1.5s. \\
Fig.~\ref{fig:joint_space_error_boxplot_ballbot} displays the boxplots of the root-mean-square error (RMSE) between the robot joint positions under each baseline and the ground-truth MPC, collecting data from 100 noise realizations affecting the measurements. {In addition, Tab.~\ref{table:final_tracking_error_ballbot} reports the final end-effector tracking error for the controllers that do not lead the arm to a singular configuration.}  \\
{As shown in Fig.~\ref{fig:joint_space_error_boxplot_ballbot}, the multi-task MPC performs best in terms of system identification in closed-loop, as it has the smallest RMSE.}
The offset compensation setup hardly manages to adapt, and indeed causes the MPC controller to generate joint-space trajectories that lead the arm to singular configurations, as it can be observed in the attached video. This is due to the fact that the induced error is state-dependent and highly non-linear, {and thus it cannot be approximated accurately with constant basis functions.} On the other hand, the single-task model tends to overfit the training trajectory, showing poor generalization capability. 
Even though the nominal MPC does not lead the arm to singularity, its performance is affected by the absence of any disturbance identification procedure, and thus also leads to a high end-effector final tracking error (see Tab.~\ref{table:final_tracking_error_ballbot}). 
In conclusion, the closed-loop state under the multi-task MPC controller is the closest to the ground-truth MPC state, improving the end-effector tracking performance.

\begin{table}[t]
\centering
\caption{Median of the final end-effector tracking error, computed over 100 simulation tests with a ballbot, as described in Sec. \ref{subsec:ballbot}.}
\setlength{\tabcolsep}{5pt}
\begin{tabular}{lrrr}
\hline
Method & \multicolumn{1}{c}{ground-truth MPC} & \multicolumn{1}{c}{multi-task MPC} & \multicolumn{1}{c}{nominal MPC} \\ \hline
Final error [m] & \textbf{0.0167}& 0.0182& 0.0765 \\ \hline
\end{tabular}
\label{table:final_tracking_error_ballbot}
\end{table}  
\subsection{Door-opening hardware experiments}
\label{subsec:door_opening_results}
\subsubsection{Experimental setup}The hardware platform employed for the experiments is the ANYmal robot with a 6-DoF arm mounted on top (Fig.~\ref{fig:title_page}).
The outputs of the MPC computations are feedforward contact forces, generalized positions and velocities, which need to be translated to joint torques. This step is performed by a QP-based whole-body controller running at 400 Hz, which additionally compensates for the full dynamics of the robot \cite{sleiman2021unified}.
All computations are executed on the robot on-board computer (Intel Core i7-
8850H CPU@4 GHz hexacore processor). The MPC time-horizon is 1 s and the MPC solution is computed at a frequency of 70 Hz. \\
In the performed experiments, the goal of the robot is to pull a door to a desired angle of 70$^{\circ}$.
The object model in Eq.~\eqref{eq:object_eom} is
$\vx_o = \begin{bmatrix} \alpha & \dot \alpha \end{bmatrix}^T$, where $\alpha, \dot \alpha$ are the door-opening angle and its derivative, respectively. The nominal flow-map $\vf_n$ assumes that the door has zero stiffness and damping. The measurement of the door dynamics error $\ve$ relies on estimating the acceleration of the door $\ddot \alpha$, which is a challenging task due to the high-impact accelerations arising during trotting \cite{sun2021online}. Here, we use a Kalman filter based on a constant acceleration model, which takes measurements of $\alpha, \dot \alpha$ as inputs, and computes estimates of $\alpha, \dot \alpha, \ddot \alpha$. This approach demonstrated to provide sufficiently smooth accelerations for the door-opening task while trotting. \\
We collect data by executing the door-opening task in the following two scenarios:
\begin{enumerate}
    \item \textbf{Added friction}: door pulling experiments with different values of friction between the door and the ground. This is realized by rigidly attaching a box to the back of the door, and {varying its weight from 0 kg to 7.11 kg.}
    \item \textbf{Human disturbance}: door pulling experiments in the presence of disturbances caused by a person.
\end{enumerate}
{The multi-task error model is trained on two datasets from the \textit{Added friction} scenario, which correspond to the added box weighing 1.6 kg and 5.35 kg, and one dataset from the \textit{Human disturbance} scenario. We point out that, in all the experiments, we use the same hyperparameters for both the frequencies of the basis functions and the Kalman filter gains, without any additional tuning.} The chosen input locations are the door state components ($\alpha$, $\dot \alpha$). 
\begin{figure}
   \centering
   \includegraphics[scale=0.52]{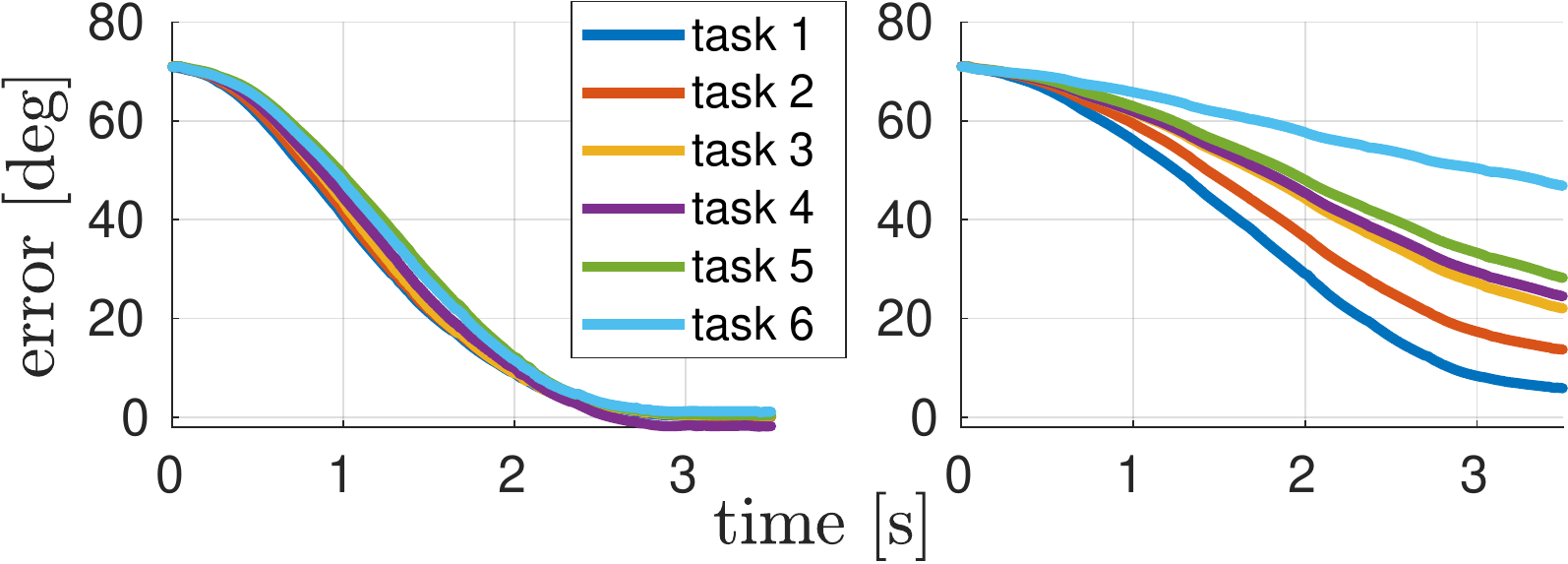}
   \caption{Tracking error for the door-opening task with added friction 
   under the multi-task MPC (on the left) and nominal MPC (on the right).} 
\label{fig:door_opening_plots_added_friction}    
\end{figure}
\begin{figure}
   \centering
   \includegraphics[scale=0.52]{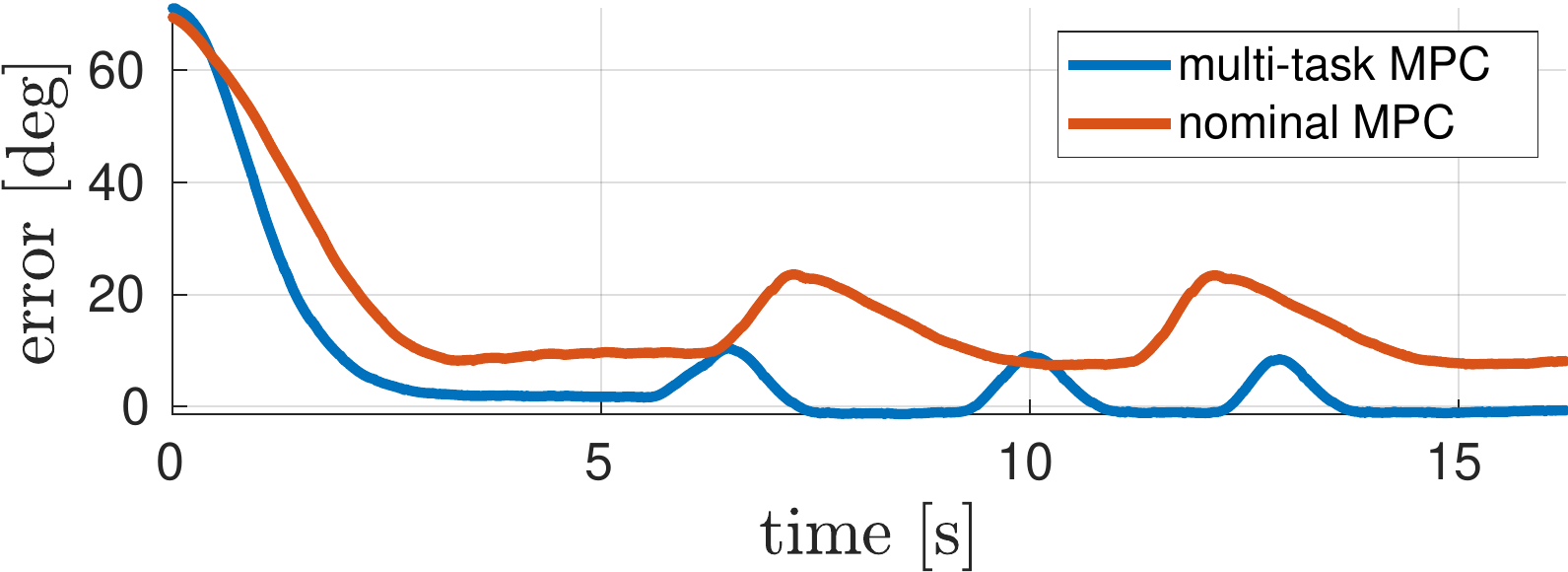}
   \caption{Tracking error under the multi-task MPC (blue curve) and nominal MPC (red curve) for a door-opening experiment in the presence of human disturbances.}
\label{fig:disturbance_rejection_door_opening}    
\end{figure}
\subsubsection{Results} 
\begin{figure}[t]
   \centering
   \includegraphics[scale=0.52]{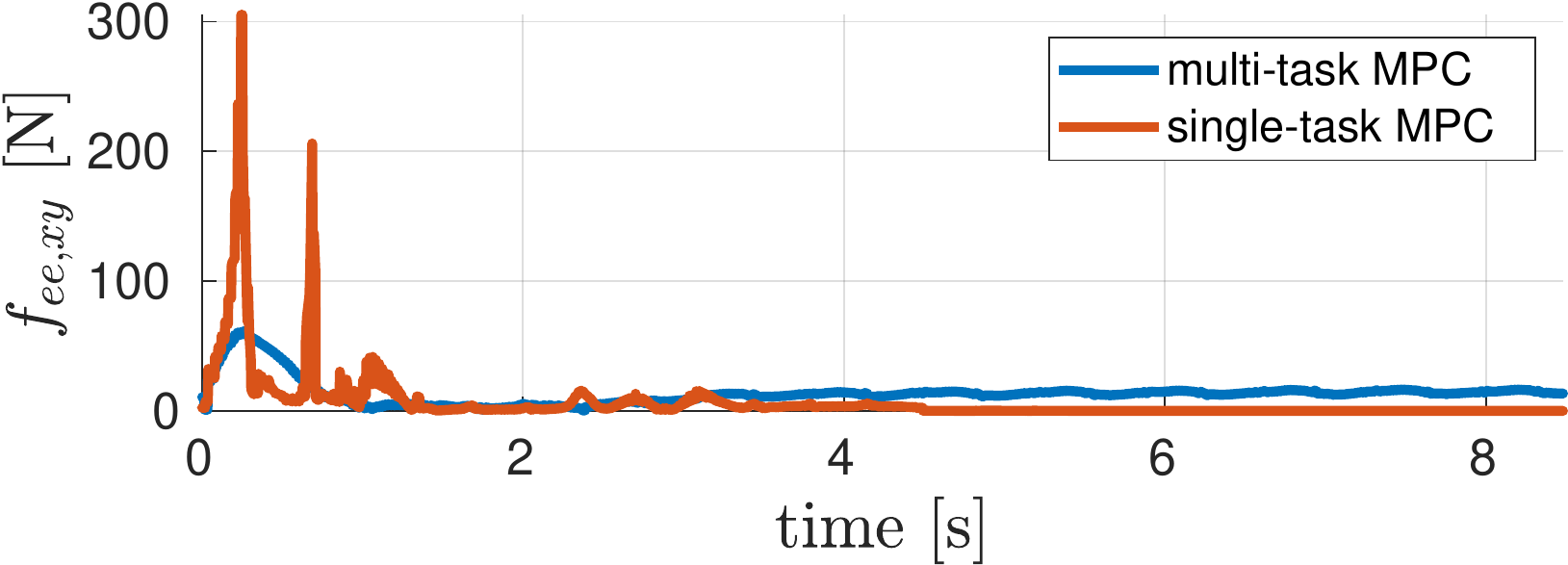}
   \caption{Norm of the planned end-effector force in a door-opening task with the multi-task MPC (blue curve) and single-task MPC (red curve).}
\label{fig:comparison_single_multi_task_input_norm}    
\end{figure}

\begin{figure}[!t]
\setlength\belowcaptionskip{-3ex}
   \centering
   \includegraphics[scale=0.52]{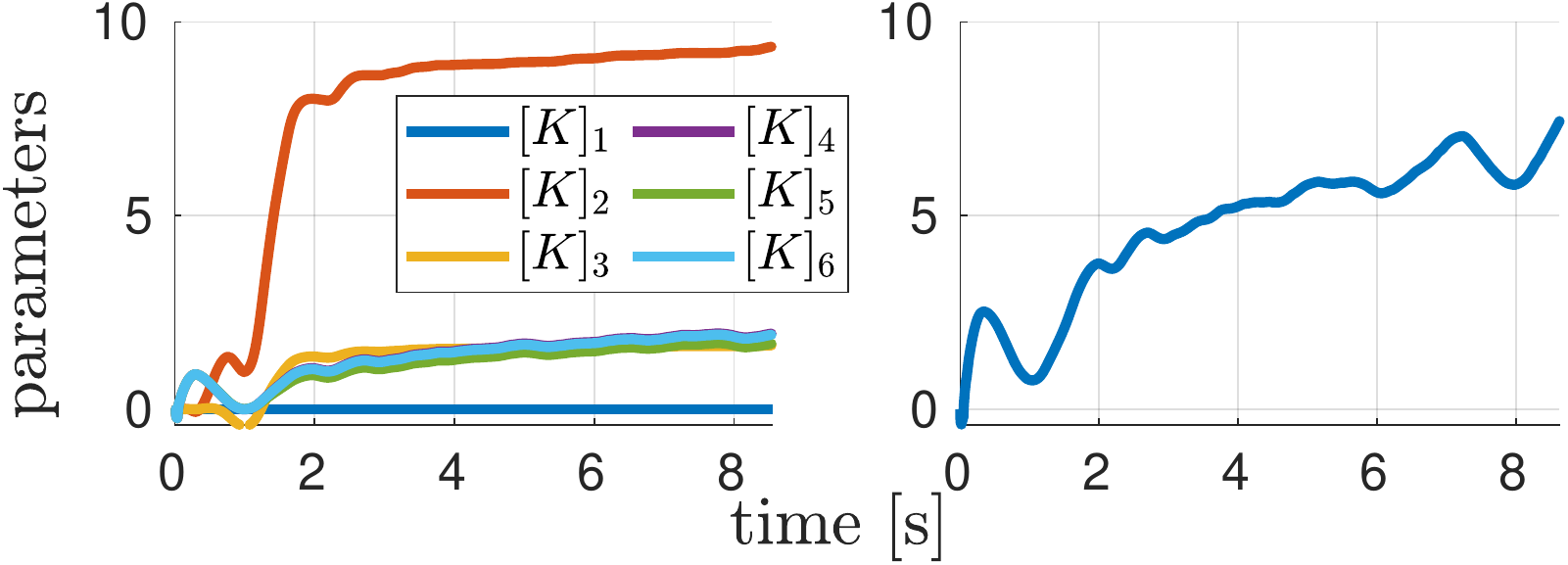}
   \caption{Task-specific parameters 
   under the multi-task MPC (on the left) and the adaptive bias of the offset compensation-based MPC controller (on the right).}
\label{fig:comparison_constant_model_parameters}    
\end{figure}

We perform 7 door-pulling experiments using the multi-task and the nominal MPC. The door angle tracking error, computed as a difference between the desired and measured door angles, is shown in Fig.~\ref{fig:door_opening_plots_added_friction} and Fig.~\ref{fig:disturbance_rejection_door_opening} for the experiments performed in the \textit{Added friction} scenario and \textit{Human disturbance} scenario, respectively. {The testing set consists of 6 tasks from the \textit{Added friction} scenario, and 1 task from the \textit{Human disturbance} scenario. These include 4 tasks, i.e. friction values, that were not seen at training time (Tasks 1, 3, 4, 6 in Fig.~\ref{fig:door_opening_plots_added_friction}).}\\ The tracking error under the nominal MPC controller shows high variability across the different tasks (Fig.~\ref{fig:door_opening_plots_added_friction}). This implies that, even if the nominal MPC controller might be well-tuned for a particular door, its performance degrades when the model of the door is perturbed or uncertain. In contrast, the tracking error under the multi-task MPC controller is consistently lower across all tasks. Similarly, in Fig.~\ref{fig:disturbance_rejection_door_opening}, the multi-task MPC correctly leads the door back to its desired position, despite repeated human disturbances. {These experiments highlight the necessity of identifying and compensating the error in the MPC system dynamics in order to have a good closed-loop tracking performance.}\\
The proposed approach is additionally evaluated against a single-task MPC controller (baseline B4), trained only on one dataset from the \textit{Human disturbance} scenario. Fig.~\ref{fig:comparison_single_multi_task_input_norm}
displays the norm of the end-effector force planned by MPC:
observing both the plot and the attached video, the behavior in closed-loop is much more aggressive under the single-task MPC compared to the multi-task MPC. This is due to oscillations present in the \textit{Human disturbance} dataset that the single-task training tends to overfit, and which do not reflect the natural spectrum of the physical system. In contrast, the multi-task MPC controller results in a smoother behavior.
The final experiment, for which no example was included in the training datasets, consists in artificially increasing the spring-stiffness of the door by adding an elastic band between the handle and the door frame. In this scenario, we compare the multi-task MPC controller against its offset compensation counterpart (baseline B3). {The results show that, in the presence of a non-constant error affecting the MPC model, compensating the uncertainty with an adaptive bias can even lead the system to instability.} Fig.~\ref{fig:comparison_constant_model_parameters} displays the task-specific parameters under the two controllers. 
%
 We observe that under the multi-task MPC, the parameters $[\vK]_1,\dots,[\vK]_6$ converge to steady-state values. In contrast, the adaptive constant bias of the offset-compensation MPC diverges. As a result, the robot tends to become unstable, as evident from the oscillations in the right plot of Fig.~\ref{fig:comparison_constant_model_parameters}, and the attached video.



\section{Conclusions}
In this work, we presented a Bayesian multi-task, learning-based MPC scheme for robotic manipulation. The core idea is to 
describe model errors as a linear combination of trigonometric basis functions based on related past experiments, enabling efficient online updates of the model for high control performance. 
We evaluated the proposed method in simulation and hardware experiments on a number of robotic scenarios, where the uncertainty in the system dynamics is either due to modeling errors in the robot model or in the manipulated object model. Our results show that the proposed approach is effective when solving control problems in the presence of uncertainties in the system dynamics, consistently outperforming the considered baselines in terms of both identification and tracking performance in closed-loop. An interesting extension to this work would be to make use of robust or stochastic techniques to provide stability guarantees for the considered model and residual uncertainties, e.g. using ideas from~\cite{JK21}. 

\section*{Acknowledgments}
We thank Daniel Mesham for supporting the implementation in the early stages of this work.

\bibliographystyle{IEEEtran}
\bibliography{IEEEexample}

\end{document}